\documentclass[conference]{IEEEtran}
\IEEEoverridecommandlockouts
\usepackage{cite}
\usepackage{amsmath,amssymb,amsfonts}
\usepackage{algorithmic}
\usepackage{graphicx}
\usepackage{textcomp}
\usepackage{xcolor}

\def\BibTeX{{\rm B\kern-.05em{\sc i\kern-.025em b}\kern-.08em
    T\kern-.1667em\lower.7ex\hbox{E}\kern-.125emX}}
\begin{document}

\title{Culture-inspired Multi-modal Color Palette Generation and Colorization: A Chinese Youth Subculture Case
}

\author{\IEEEauthorblockN{Yufan Li}
\IEEEauthorblockA{
\textit{Tongji University}\\
Shanghai, China\\
1831982@tongji.edu.cn}
\and
\IEEEauthorblockN{Jinggang Zhuo}
\IEEEauthorblockA{
\textit{Tongji University}\\
Shanghai, China\\
zhuojg1519@tongji.edu.cn}
\and
\IEEEauthorblockN{Ling Fan}
\IEEEauthorblockA{
\textit{Tongji University; Tezign.com Ltd.}\\
Shanghai, China\\
lfan@tongji.edu.cn}
\and
\IEEEauthorblockN{Harry Jiannan Wang}
\IEEEauthorblockA{
\textit{University of Delaware}\\
Newark, DE, USA\\
hjwang@udel.edu}
}

\maketitle

\begin{abstract}
Color is an essential component of graphic design, acting not only as a visual factor but also carrying cultural implications. However, existing research on algorithmic color palette generation and colorization largely ignores the cultural aspect. In this paper, we contribute to this line of research by first constructing a unique color dataset inspired by a specific culture, i.e., Chinese Youth Subculture (CYS), which is an vibrant and trending cultural group especially for the Gen Z population. We show that the colors used in CYS have special aesthetic and semantic characteristics that are different from generic color theory. We then develop an interactive multi-modal generative framework to create CYS-styled color palettes, which can be used to put a CYS twist on images using our automatic colorization model. Our framework is illustrated via a demo system designed with the human-in-the-loop principle that constantly provides feedback to our algorithms. User studies are also conducted to evaluate our generation results.
\end{abstract}


\section{Introduction}
\begin{figure}[htbp]
\centerline{\includegraphics[width=0.5\textwidth]{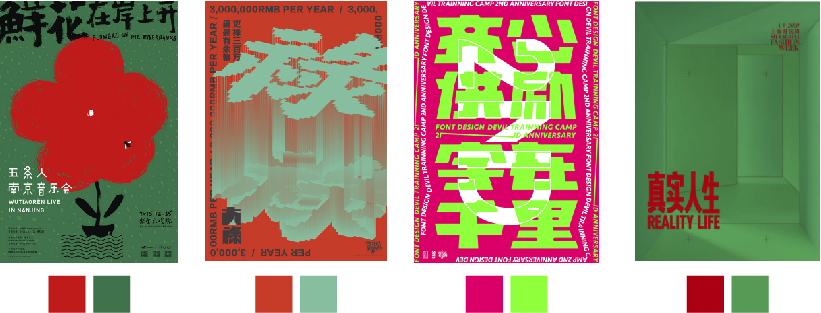}}
\caption{Examples of ‘red with green’ designs in Chinese youth subculture}
\label{fig1}
\end{figure}

Color is an essential component of graphic design, acting as a visual factor while also conveying human inner emotions. These emotions and impacts are strongly influenced by the cultural implications behind colors. For example, the combination of red and green is unpleasant according to traditional Chinese color theory but this same color combination represents a cool and rebellious style for Chinese Youth Subculture (CYS) groups as illustrated by a number of posters found in popular CYS websites (Fig.~\ref{fig1}). In addition, colors may have different semantic meanings for different cultures. For instance, the color directly related to money is green in Western culture, which is often gold and red in Eastern culture. Therefore, culture is an important factor that cannot be ignored when studying color semantics.

\begin{figure*}[htbp]
\centerline{\includegraphics[width=0.95\textwidth]{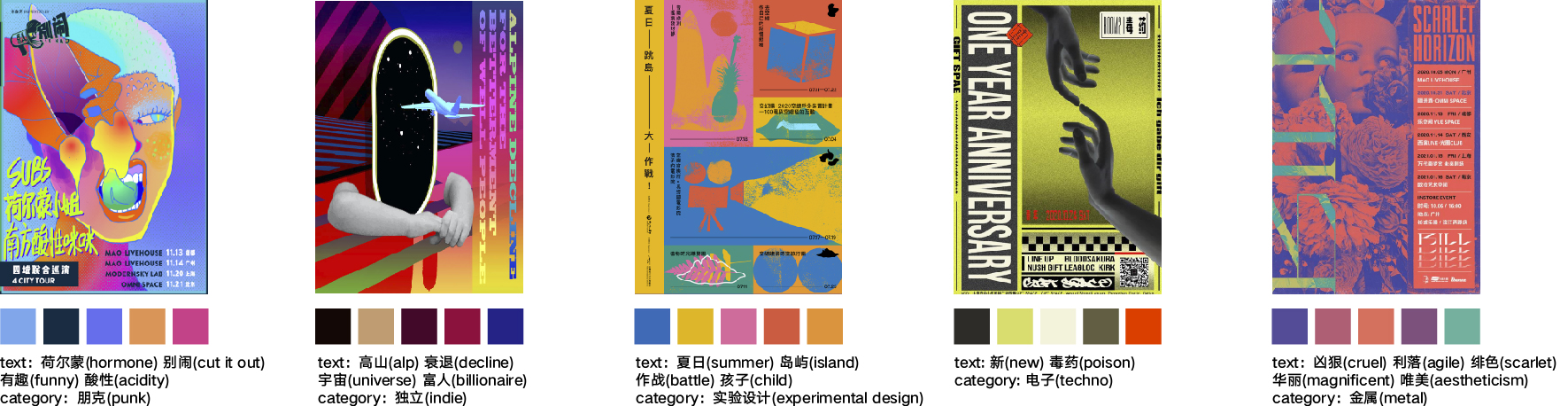}}
\caption{CYS dataset examples}
\label{fig2}
\end{figure*}

Color has also been an active research topic in the machine learning and AI areas\cite{b11}\cite{b12}\cite{b13}\cite{b14}. Many design tool companies have intelligent color design as one of their main product features, such as the Adobe Color tool by Adobe (color.adobe.com) and the Color Match tool by AI startup FacetAI (facet.ai). However, existing researches and tools on algorithmic color palette generation and colorization \cite{b10}\cite{b15}  largely ignore the cultural aspect. In this paper, we aim to extend existing work by incorporating the culture aspect. Created by and disseminated among Chinese young population, Chinese Youth Subculture has the unique balance of confliction and cooperation with the mainstream culture and its parent culture, symbolizing the identity of the young people in China \cite{b1}. We want to focus on the color research of the Chinese Youth Subculture for a number of reasons. First, based on our interviews of a number of graphic designers, who have done projects related to CYS, the colors used in CYS have special aesthetic and semantic characteristics that are different from existing color theories, which calls for more research. Second, previous research\cite{b2} has proved that the integration of subcultural style into the mainstream style is the driving force of design innovation, which keeps the mainstream culture alive and fresh. Finally, CYS is an vibrant and trending cultural group especially for the Gen Z population, which is one of the main driving forces of the 21st century economy and is of great interest to many businesses. Therefore, our study of CYS color system can have broader impact and practical design and business implications.

Besides color palette, there are other important factors that can affect the overall CYS style of the images, such as image elements, composition, text, and context. Take Fig.~\ref{fig1} as an example, with a similar palette of red with green, the first poster has a red flower on a green background while the second poster has a green Chinese character on a red background, indicating different roles of elements and composition. The text in the image also has effect on the color, e.g., the text of the first poster is ``flower blossoms on a river bank'', which implies the flower is a blossoming bright red flower on a fresh green bank. There is also additional text description of the poster that influences the CYS style design as we discuss in our dataset later. In addition, the context of the image also correlate with the CYS style. For instance, the first poster is for a live show of a rock band, which is casual and entertaining, while the fourth poster that is a commercial poster aimed at CYS-groups, which is more rigid and serious. 

In this paper, we address the aforementioned CYS color research challenges by proposing an open-source culture-inspired multi-modal color palette generation and colorization framework. Our contributions are three-fold:

\begin{itemize}
\item We construct a unique open dataset of images, color palettes, text and categories for CYS to facilitate culture related color research.
\item We introduce an innovative multi-modal generative architecture for culture-inspired color palette generation and colorization. 
\item We develop a demo system to illustrate our framework, which incorporates the human-in-the-loop principle to enable automatic system evolution.
\end{itemize}

The dataset, code, and demo have been open sourced. \footnote{https://github.com/tezignlab/subculture-colorization}

\section{Related Work}

Color and culture is an important branch in the study of art. The book ``The Theory of Color and Culture'' \cite{b3} comprehensively expounds on the relationship between color and culture. Sharkhuu et al. studied the use of colors in Buddhist cultures in different countries based on Buddhist pictures\cite{b4}. Kim et al. paid attention to the color of photos posted on the social platform Instagram and used colorfulness, color diversity, and color harmony to study the characteristics of different user groups\cite{b5}. Zhe et al. used the K-Means clustering method to extract the typical colors of traditional Chinese embroidery and studied the color characteristics of traditional Chinese cultural phenomena\cite{b6}. No study has made further algorithmic color generation attempt for a specific culture.

Color is not only an objective physical attribute but also strongly arouses people’s semantic associations. Koshiyaba\cite{b7} was the first to quantify the color-semantic relationship by using a color semantic space containing 180 adjectives expressing 1170 3-color palettes. Bahng et al.\cite{b10} proposed a GAN-based model to generate color palette based on a dataset with 4,312 words and 10183 text-palette pairs. In addition, many practical color semantic online tools have appeared, such as Adobe Color and Color Hex. Researchers have also used different interactive coloring methods for gray-scale images, such as reference images method\cite{b11}\cite{b12} and users’ stroke hints approach\cite{b13}. Text-based colorization has also been studied \cite{b10}\cite{b15}. However, no studies have considered the impact of culture on the text-palette relationship. 

\begin{figure}[htbp]
\centerline{\includegraphics[width=0.48\textwidth]{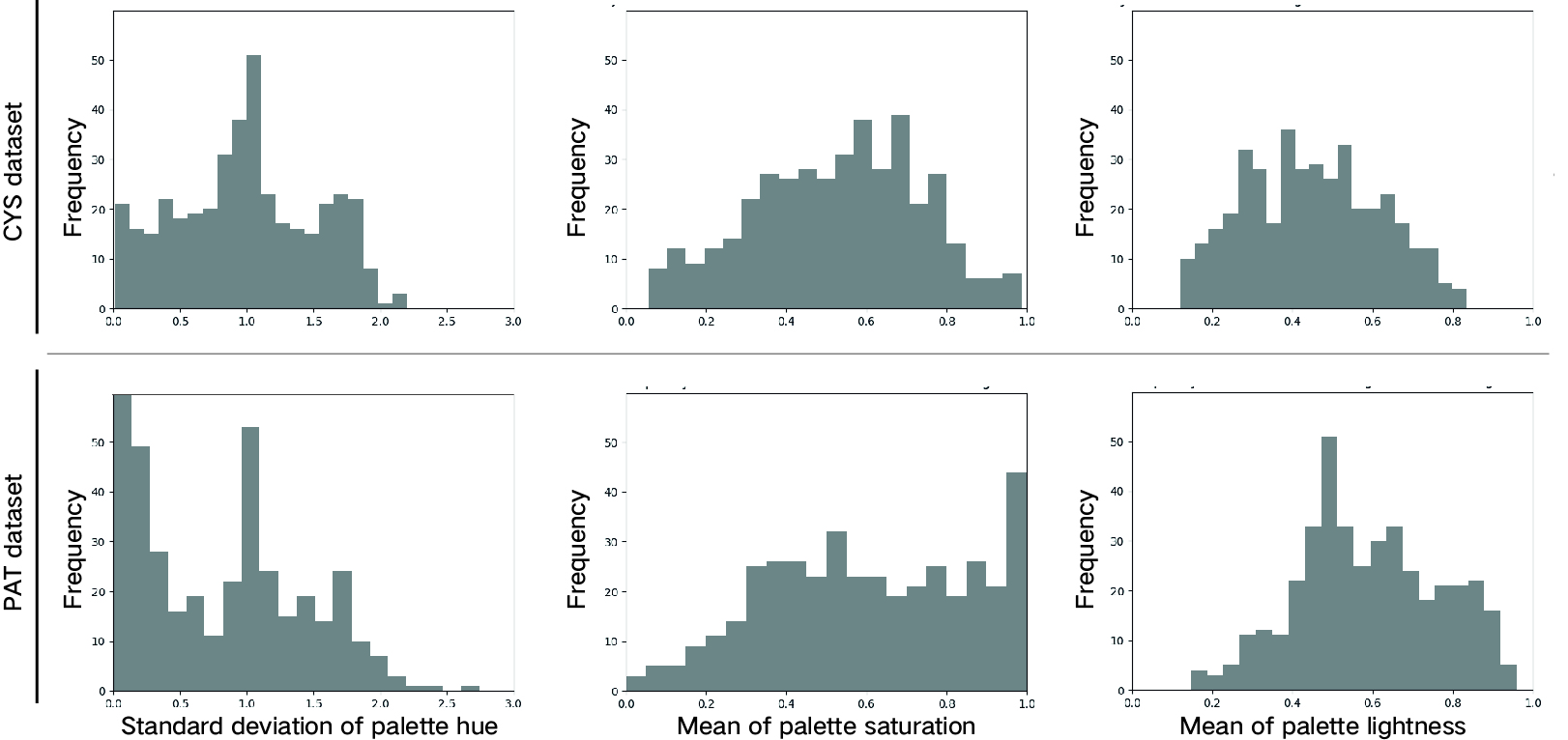}}
\caption{Comparison between CYS dataset and PAT dataset}
\label{fig3}
\end{figure}

\begin{figure*}[htbp]
\centerline{\includegraphics[width=0.95\textwidth]{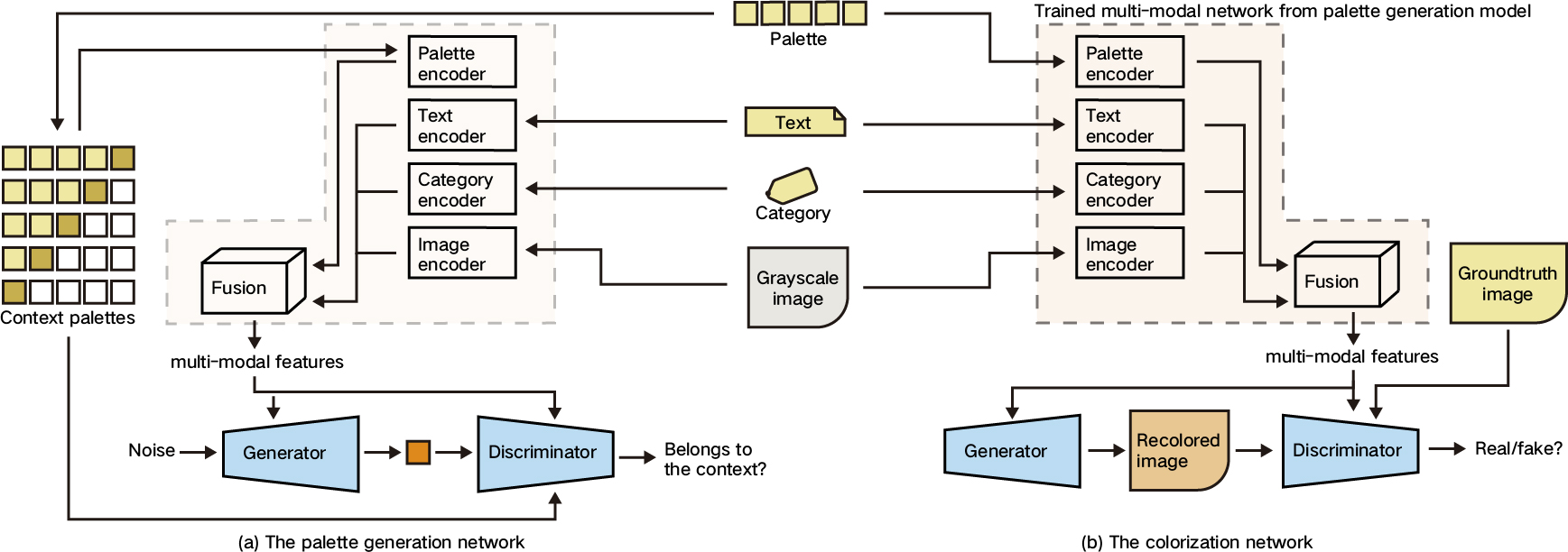}}
\caption{The training structure of our framework}
\label{fig4}
\end{figure*}

\section{CYS Color Dataset}
We manually constructed the CYS color dataset, which contains 1263 images with corresponding 5-color palette, descriptive text, and category. Fig.~\ref{fig2} shows five examples from the dataset and the dataset development process is discussed next. Based on the interviews of designers who have done CYS-related projects, we obtain a list of websites targeting CYS population and manually collected 7051 images from those website, which are posters, album covers, and illustrations. Then, three designers manually selected 1263 images that have strong CYS style based on their judgement. From the web pages where the images are collected, we synthesized a list of keywords and phrases related to the CYS expression of the images by removing text irrelevant to color and CYS, resulting in 2535 unique Chinese adjectives, nouns, and verbs. We also collected the categories data from the web pages, includes 14 categories, such as punk, hiphop, techno, etc. For each image, we first extract 10 colors using existing clustering based algorithm and then let designer select 5 colors to form the color palette according to the following rules: 1) removing similar colors that cannot be easily separated, 2) selecting the color combination that best represents the visual effect according to image elements and composition, and 3) selecting the colors that best reflects CYS and rank them accordingly. This step is critical because the clustering algorithm essentially ranks the colors based on their proportions in the images, which may miss many important colors for CYS style if we only choose the first 5 colors.  

In order to show that CYS dataset is indeed different from generic color datasets, we conducted a comparison with the PAT dataset from Bahng et al.\cite{b10}, which does not consider the cultural aspect. We randomly sampled 400 images from each dataset and calculated the overall distributions of the HSL (hue, saturation, lightness) values as shown in Fig.~\ref{fig3}. The plot and T test results \footnote{p-values for hue standard deviation, lightness mean, and saturation mean are $1.11*10^{- 7}, 1.93*10^{- 6}, 2.42*10^{- 29}$ respectively.} clearly show the differences, e.g., CYS colors tend to have more diverse hue and lower saturation and lightness.

\section{Framework and Demo}
As shown in Fig.~\ref{fig4}, our framework includes two separately trained networks: the color palette generation network and the colorization network. The first network is trained through a conditional GAN (cGAN) with a multi-modal input to generate CYS color palettes. Another cGAN in the second model is trained to color the input images according to the palette generated by the first network. 

In order to augment the color dataset to provide more training samples, we leverage the relationships among the colors in the palette and ask the discriminator to determine whether the generated color is considered the next color in the palette. Previous colors are fused together by a fusion function with the other three inputs to form the multi-modal context, which is fed to the cGAN as a condition to predict the next color. Finally, 5 generated colors are concatenated and output as the color palette.

We assign different weights to different elements in the multi-modality module using a parameter $\lambda$ ($\lambda_{text}=0.5$, $\lambda_{image}=0.4$, $\lambda_{category}=0.1$ are default settings). The corresponding encoders are $E_{image}$, $E_{text}$ and $E_{category}$ and the combined context of grayscale image, text and category is defined as $c_1 = \sum_{c}^{}\lambda_cE_c(x_c), c\in\{image, text, category\}$. Given the encoded palette $ c_2 = E_{palette}(palette) $ and final multi-modal context is defined as $ y=f(c_1, c_2) $. We follow the least-squares GAN (LSGAN) \cite{b16} to formulate the loss functions as:

$L_D=\alpha(D(x, y)-1)^2+(1-\alpha)(D(G(z, y), y))^2$

$L_G=\alpha(D(G(z, y), y)-1)^2$

where $\alpha=0.5$ is hyperparameter, $x$ is the ground truth, $y$ is the multi-modal context, and $z$ is random noise.

The colorization network is structurally similar to the palette generation network, but use the generated 5-color palette as the input and different set of weights for the multi-modality context ($\lambda_{text}=0.3$, $\lambda_{image}=0.6$, and  $\lambda_{category}=0.1$).

\begin{figure}[htbp]
\centerline{\includegraphics[width=0.5\textwidth]{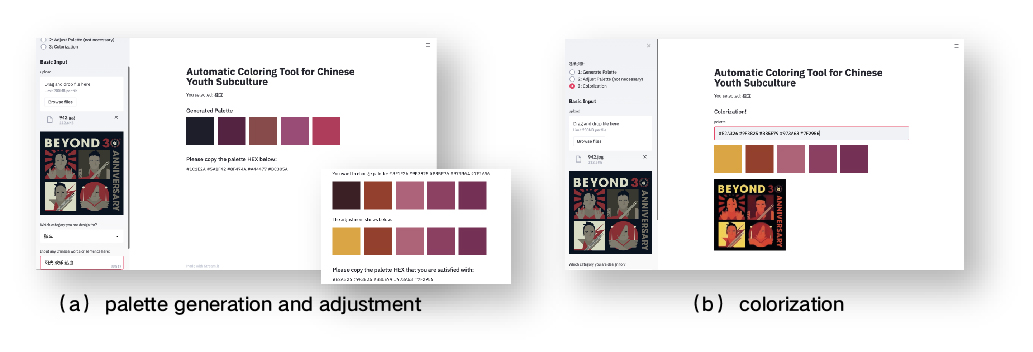}}
\caption{The demo system}
\label{fig5}
\end{figure}

As shown in Fig.~\ref{fig5}, we have developed a demo system to materialize our framework\footnote{video of the demo: https://youtu.be/fLPwr-oX0ds}, where users can obtain a image that is colored with the CYS style by following three steps: 1) After entering the text description, uploading the grayscale image and selecting the category, the user can get the generated CYS-styled palette, 2) the color palette can be adjusted according to the user's preference, which is recorded by the system to improve the algorithm, 3) the image uploaded in the first step will be colorized by our system to add CYS twist. 

\begin{figure}[htbp]
\centerline{\includegraphics[width=0.5\textwidth]{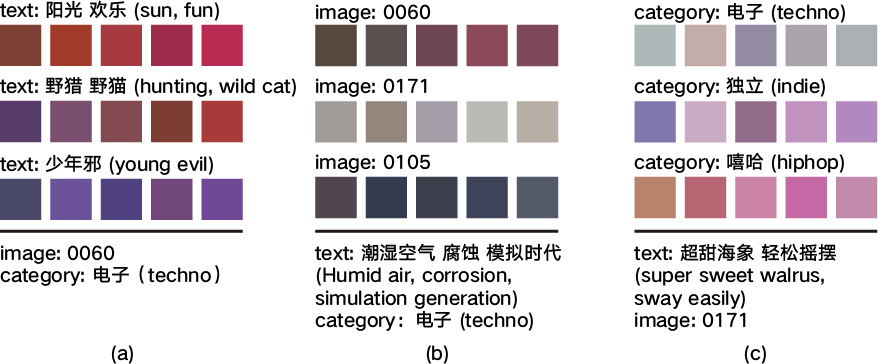}}
\caption{Evaluation of multi-modility (a) different text with controlled image and category; (b) different image with controlled text and category; (c) different category with controlled text and image.}
\label{fig6}
\end{figure}

\section{Evaluation}

We want to evaluate our framework from two perspectives, i.e., whether the multi-modal context affects the generation results, and whether the colorized images have CYS styles. 

As shown in Fig.~\ref{fig6}, we did three sets of experiments for the first perspective: a) changing text while fixing image and category, b) changing image while fixing text and category, and c) changing category whiling fixing image and text. The results show that each of the multi-modal context components has direct impact on the generation results and our generation model has excellent controlled diversity. For example, in Fig.~\ref{fig6} (a), after changing the text from ``sun, fun'' to ``evil'', the generated colors are apparently darker to reflect effect of the semantic textual changes.

\begin{figure}[htbp]
\centerline{\includegraphics[width=0.5\textwidth]{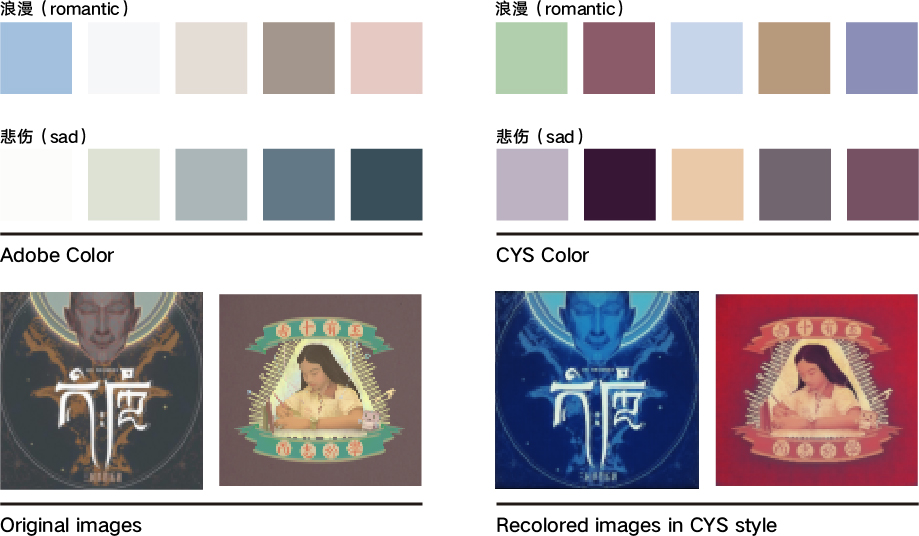}}
\caption{User study examples}
\label{fig7}
\end{figure}

We design a user study to evaluate from the second perspective. We choose 20 keywords and search the Adobe Color website to get the cultural neutral color palettes. Then, we generate our color palette using the keywords, 20 cultural neutral images that are from a graphic design website and category ``indie'' to form the color palette pairs. Then, we use the generated color palettes to colorize 20 images and form the image pairs. The palette pairs and image pairs are shown in Fig.~\ref{fig7}. We then ask four designers to review the color palette and image pairs and determine which one is more CYS styled. The results show that 75\% of our palettes and 76\% of recolored images are considered to be more in line with the CYS-style, which verifies that our framework learned the color specificity of CYS.

\section{Conclusion and Future Work}
In this paper, we proposed a color generation and colorization framework based on a manually collected Chinese Youth Subculture dataset. A demo system of the framework had been developed and evaluated via a set of experiments and a user study. To the best of our knowledge, our framework is the first culture-inspired approach to automatic color generation and colorization. Note that our framework is not limited to CYS culture and can be used for any culture if trained using culture specific data. We are in the process of expending the dataset and extending our framework by calibrating additional design features that may have important influence to culture-based color theory.

\vspace{12pt}
\color{red}

\end{document}